\documentclass[11pt,a4paper]{article}
\usepackage[draft]{hyperref}
\usepackage[hyperref]{acl2018}
\usepackage{times}
\usepackage{latexsym}

\usepackage{epsfig} % for postscript graphics files
\usepackage{amsmath} % assumes amsmath package installed
\usepackage{amssymb}  % assumes amsmath package installed
\usepackage[english]{babel}
\usepackage{multirow}
\usepackage{booktabs,caption}
\usepackage{graphicx}
\usepackage{multicol}
\usepackage{tabularx}
\usepackage[utf8]{inputenc}
\usepackage{xspace}
\usepackage{booktabs}
\usepackage{relsize}

\newcommand{\tabref}[1]{Table~\ref{#1}\xspace}

\newcommand{\DIFFVEC}[1][]{\textsc{DiffVec#1}\xspace}
\newcommand{\DOCDIFFVEC}[1][]{\textsc{DocDiffVec#1}\xspace}

\newcommand{\model}[1]{\textsc{#1}\xspace}
\newcommand{\WR}{\model{WR}}
\newcommand{\Average}{\model{Average}}

\newcommand{\method}[1]{\textsf{#1}\xspace}
\newcommand{\wordtovec}{\method{word2vec}}
\newcommand{\paragram}{\method{Paragram}}
\newcommand{\glove}{\method{GloVe}}
\newcommand{\infersent}{\method{Infersent}}
\newcommand{\bilstmavg}{\method{BiLSTM-NMT$_{\text{avg}}$}}
\newcommand{\bilstmmax}{\method{BiLSTM-NMT$_{\text{max}}$}}
\newcommand{\bilstm}{\method{BiLSTM-NMT}}
\newcommand{\doctovec}{\method{doc2vec}}
\newcommand{\dbow}{\method{dbow}}
\newcommand{\cbow}{\method{cbow}}
\newcommand{\skipgram}{\method{skip-gram}}
\newcommand{\crf}{\method{CRF}}
\newcommand{\svmhmm}{\method{SVM-HMM}}
\newcommand{\transr}{\method{TransR}}
\newcommand{\crfsgd}{\method{CRFSGD}}
\newcommand{\me}{\method{ME}}
\newcommand{\svm}{\method{SVM}}
\newcommand{\skipthought}{\method{SkipThought}}
\newcommand{\fastsent}{\method{FastSent}}

\newcommand{\dataset}[1]{\texttt{#1}\xspace}
\newcommand{\cqadupstack}{\dataset{cqadupstack}}
\newcommand{\cnet}{\dataset{cnet}}
\newcommand{\wiki}{\dataset{WIKI}}
\newcommand{\AP}{\dataset{AP}}

\newcommand{\lex}[1]{\textit{#1}\xspace}

\newcommand{\transp}{\ensuremath{\intercal}}
\renewcommand{\vec}[1]{\ensuremath{\mathbf{#1}}\xspace}

\newcommand{\da}[1]{\textsc{#1}\xspace}
\newcommand{\Qq}{\da{Question-Question}}
\newcommand{\Qc}{\da{Question-Correction}}
\newcommand{\Aa}{\da{Answer-Answer}}
\newcommand{\Rr}{\da{Resolution}}

\title{Evaluating the Utility of Document Embedding Vector Difference for Relation Learning}
\author{Jingyuan Zhang \\ jingyuanz@student.unimelb.edu.au \And
               Timothy Baldwin \\ tb@ldwin.net}

\begin{document}

\maketitle

%%%%%%%%%%%%%%%%%%%%%%%%%%%%%%%%%%%%%%%%%%%%%%%%%%%%%%%%%%%%%%%%%%%%%%%%%%%%%%%%
\begin{abstract}
  Recent work has demonstrated that vector offsets obtained by
  subtracting pretrained word embedding vectors can be used to predict lexical relations with surprising
  accuracy. Inspired by this finding, in this paper, we extend the idea
  to the document level, in generating document-level embeddings,
  calculating the distance between them, and
  using a linear classifier to classify the relation between the
  documents. In the context of duplicate detection and dialogue act
  tagging tasks, we show that document-level difference vectors have utility in assessing
  document-level similarity, but perform less well in multi-relational
  classification.
\end{abstract}

%%%%%%%%%%%%%%%%%%%%%%%%%%%%%%%%%%%%%%%%%%%%%%%%%%%%%%%%%%%%%%%%%%%%%%%%%%%%%%%%
\section{Introduction}

Document-level relation learning has long played a significant role in Natural Language Processing (``NLP''), in tasks including semantic textual similarity (``STS''), natural language inference, question answering, and link prediction.  

Word and document embeddings have become ubiquitous in NLP, whereby words or documents are mapped to vectors of real numbers. Building off this, the work of \newcite{mikolov2013distributed} and \newcite{P16-1158} demonstrated the ability of word embedding offsets (``\DIFFVEC[s]'') in completing word analogies (e.g.\ A:B :: C:-?-). For example, the \DIFFVEC between \lex{king} and \lex{queen} is roughly identical to that between \lex{man} and \lex{woman}, indicating that the \DIFFVEC may imply a relation of OPPOSITE-GENDER in its magnitude and direction, which offers a support of analogy prediction tasks of the form (\lex{king}:\lex{queen} :: \lex{man}:-?-). In this paper, we evaluate the utility of document embedding methods in solving analogies in terms of document relation prediction. That is, we evaluate the task of \textit{document} embedding difference (``DOCDIFFVEC'') to model document relations, in the context of two tasks: document duplication detection, and post-level dialogue act tagging. In doing so, we perform a contrastive evaluation of off-the-shelf document embedding models.

We select a range of document embedding methods that are trained in either an unsupervised or supervised manner and have been reported in recent work to perform well across a range of NLP transfer tasks. In line with \newcite{P16-1158}, we keep the classifier set-up used to perform the relation classification deliberately simple, in applying a simple linear-kernel support vector machine (``SVM'') to the \DOCDIFFVEC[s]. Our results show that \DOCDIFFVEC has remarkable utility in binary classification tasks regardless of the simplicity of the model. However,  for multi-relational classification tasks, it only marginally surpasses the baseline model, and unsupervised averaging models are superior to more complex supervised models.

\section{Related Work}

Recently, advanced word embeddings and neural network architectures have been increasingly used to model word sequences, achieving impressive results in contexts including machine translation, text classification, and sentiment analysis.

\subsection{Word Embeddings}

In work that revolutionised NLP, \newcite{mikolov2013distributed} proposed word2vec as a means of ``pre-training'' word embeddings from a large unannotated corpus of text, based on language modelling, i.e.\ predicting contexts from a word or words from context. Subsequently, others have proposed methods, including GloVe \cite{pennington2014glove} and Paragram \cite{Q15-1025}.

word2vec (``\wordtovec'') is a predict-based model, in the form of either the \skipgram or \cbow (continuous Bag-Of-Words) model. The \skipgram model aims to predict context from a target word, while \cbow, predicts the target word from its context words. The general idea can be explained as follows: given a predicted word vector $\vec{\hat{r}}$ and a target word vector $\vec{w_t}$, the probability of the target word conditional on the predicted word is calculated by a softmax function:
\begin{equation*}
  P(\vec{w_t} | \vec{\hat{r}}) = \frac{\exp(\vec{w_t}^\transp\vec{\hat{r}})}{\sum_{w \in W} \exp(\vec{w}^\transp\vec{\hat{r}}) }  
\end{equation*}
where $W$ is the set of all target word vectors. \wordtovec is trained to minimise the negative log-likelihood of the target word vector given its corresponding predicted word.

\glove performs a low-rank decomposition of the corpus co-occurrence frequency matrix based on the following objective function:
\begin{equation*}
  J = \frac{1}{2} \sum_{i, j=1}^{V} f(P_{ij})(\vec{w_i}^\transp \tilde{\vec{w_j}} - \log P_{ij})^2
\end{equation*}
where $\vec{w_i}$ is a vector for the left context, $\vec{w_j}$ is a
vector for the right context, $P_{ij}$ is the relative frequency of word $j$ in the context of word $i$, and $f$ is a heuristic weighting function to balance the influence of high versus low term frequencies.

Paragram (``\paragram'') is trained with supervision over the Paraphrase Database (PPDB) \cite{ganitkevitch2013paraphrase}, such that embeddings for expressions which are paraphrases of one another have high cosine similarity, and non-paraphrase pairs have low similarity.  The embedding for an expression is generated by simple averaging over the word embeddings, with the ultimate result of training the model  being word embeddings.

\subsection{Document Embeddings}

Recently, a lot of work has focused on obtaining ``universal'' representations for documents. Such models vary vastly in complexity and training approaches. Two typical categories of training methodologies are unsupervised and supervised. Unsupervised document embedding methods like \skipthought\cite{conf/nips/KirosZSZUTF15} and \fastsent\cite{N16-1162} are trained without supervision using neural networks on large corpora. Some simpler ones are based on pure arithmetic operations over word embedding vectors without training. Such models include averaging, weighted averaging, and weighted averaging with PCA projection \cite{arora2017asimple} (``\WR'', hereafter). The \WR model performs weighted averaging over word vectors according to word frequencies in the corpus, and adds an additional layer which modifies the final representation of the sentences using PCA projection. 

On the other hand, supervised models are based on richer compositional architectures including deep feed-forward neural networks \cite{P15-1162}, convolutional neural networks \cite{D14-1181}, attention-based networks \cite{N16-1174}, and bidirectional recurrent neural networks, among which the most popular approaches use Bi-LSTMs \cite{journals/taslp/PalangiDSGHCSW16,D15-1167,journals/corr/WietingBGL15a}.
For a sentence consisting of $T$ words $\{\vec{w_{t}}\}_{t=1,...,T}$, a bi-directional LSTM computes a set of $T$ vectors $\{\vec{h_{t}}\}_{t}$ for $t \in [1,....,T]$, formed through the concatenation of a forward and backward LSTM \cite{D17-1070}.
To combine values from each dimension of the LSTM hidden states, two common pooling methods are max pooling \cite{collobert2008unified} and average pooling.

Such models require training data, generally in the form of paraphrastic text corpora or inference datasets. 

\infersent \cite{D17-1070} is one state-of-the-art supervised method, which is trained over natural language inference datasets SNLI \cite{D15-1075} and MultiNLI \cite{DBLP:conf/naacl/WilliamsNB18}. The authors suggest a BiLSTM with max pooling as the best configuration for \infersent, and provide a pre-trained model which generates 4096-dimensional document embeddings.

Another method proposed by \newcite{DBLP:conf/acl/GimpelW18} is also based on a BiLSTM architecture with either average pooling or max pooling, but trained on a back-translated corpus called PARANMT-50M, containing over 51 million English--English sentential paraphrase pairs based on CzEng 1.6 \cite{conf/tsd/BojarDKLNPSV16}. We refer to this model as ``\bilstmavg'' or ``\bilstmmax'', for the average and max pooling methods, respectively.

\section{Methodology and Resources}

\subsection{Relation Learning}

Lexical relations for words take the form of a directed binary relation between a word pair. For example, (\lex{take}, \lex{took}) has the relation of PAST-TENSE, and (\lex{person}, \lex{people}) indicates the PLURAL relation. Recent approaches to lexical relation learning based on representation learning have made noteworthy contributions to NLP tasks, including relation extraction, relation classification, and relation analogy \cite{P16-1158}. A considerable amount of research has been dedicated to finding an explanation for the success of word embedding models in lexical relation learning, including those that focused on vector differences (\DIFFVEC). Research by \newcite{P16-1158} was the first to systematically test both the effectiveness and generalizability of \DIFFVEC across a broad range of lexical relations. The authors performed both clustering and classification experiments on a dataset consisting of over 12,000 $(relation, word_1, word_2)$ triples covering 15 relation types. Clustering of \DIFFVEC[s] revealed that many relations formed tight clusters with clear boundaries. Based on this finding, they trained supervised models on \DIFFVEC[s] to reveal their true potential in classifying lexical relations. With the addition of negative sampling, they were able to achieve impressive performance in capturing semantic and syntactic differences between words using a simple \svm model trained on \DIFFVEC[s] for both open- and closed-world classification experiments. The paper also conducts cross-comparison of different methods for learning word embedding vectors to generate \DIFFVEC[s].

Correspondingly, document relations are relations between document pairs. Document relations can also be described in the form of binary relations, similar to word-level ones. For example, (\lex{I walk to school}, \lex{I go to school on foot}) can be viewed as having a synonym or paraphrastic relation. With research having shifted focus from word embeddings to document embeddings, more efforts have been put into learning document relations. In recent years, an increasing number of shared datasets have been generated to improve certain types of sentential relation learning, including SemEval \cite{journals/corr/abs-1708-00055} and the work of \citet{Lee05anempirical} for STS; SNLI \cite{D15-1075} and MultiNLI \cite{DBLP:conf/naacl/WilliamsNB18} for natural language inference; and WikiQA \cite{conf/emnlp/YangYM15} and QAsent \cite{conf/emnlp/WangSM07} for question-answering.

The majority of models reported to perform well on document relation learning tasks depend on a joint model to aggregate two sentence vectors in a relation tuple for relation representation. This includes the joint model from \infersent that learns entailment relations, CNN-based joint models for learning paraphrastic and question-answering relations \cite{N15-1091}, and the widely used cosine similarity between sentence vectors. However, no systematic evaluation has been performed on using document vector offsets (\DOCDIFFVEC) alone for document relation learning. Taking our lead from \DIFFVEC, this paper is the first to naturally extend the \DIFFVEC evaluation paradigm to the document level.

\subsection{Learning Scheme}

For document relation learning, an aggregation mechanism is required to combine sentence vectors $\vec{h_1}, \vec{h_2}$ into a single relation vector $\vec{h_r}$. In this paper, the aggregation model is as simple as calculating $\vec{h_r}$ by subtracting $\vec{h_2}$ from $\vec{h_1}$ in a given $(\vec{h_1}, \vec{h_2}, r)$ triple, with relation $r$. That is, classification is performed over instances of form $(\vec{h_1} - \vec{h_2}, r)$.

The objective is to train a model with all $(\vec{h_1} - \vec{h_2}, r)$ training instances, and best predict the missing $r$ from within the task domain for test relation tuples $(\DOCDIFFVEC, -?-)$. Following \citet{P16-1158}, the learner in our experiments is a linear kernel SVM.

In this paper, we assess the utility of \DOCDIFFVEC[s] in learning document relations in two scenarios: (1) document-level similarity modelling, and (2) multi-relational classification.

In a document-level similarity modelling context over unordered document pairs, there is no well-defined way of ordering the documents to perform the vector difference when calculating the vector offsets. To take an (overly) simplified example, for the sentence pair (\lex{The man put the box down}, \lex{The man dropped the box}) encoded into the 3-d vector pairing $((1, 1, 0), (1,0,0.2))$, and the unordered lexical relation of paraphrastic similarity, it is impossible to define a priori which of these two sentences should be the subtrahend or the minuend. Directly taking the offset will result in two possible \DOCDIFFVEC[s] $(0,1,-0.2)$ or $(0,-1,0.2)$ depending on the ordering of the two sentences. We make the simplifying assumption that for similarity modelling, the relation only depends on the magnitude and not the direction of the \DOCDIFFVEC, and therefore calculate $\DOCDIFFVEC[s] = |\vec{h_1} - \vec{h_2}|$ using the element-wise absolute value for the offsets, eliminating the impact of directionality in each dimension.

\section{Datasets}

\subsection{CQAdupstack Dataset}

The \cqadupstack dataset \cite{conf/adcs/HoogeveenVB15} is focused on the tasks of question-answering and thread duplicate detection. In this paper, we only consider the thread duplication detection setting (``Q-DUP'' hereafter). The dataset consists of question threads crawled from 12 StackExchange\footnote{\url{https://stackexchange.com/}} subforums. Nowadays, with the number of questions asked on Q\&A forums growing dramatically, many newly-posted questions overlap in content with previously posted (and answered) questions. Q-DUP is the task of automatically identifying questions which pre-exist in the forum, and prompting the question-answerer with possible solutions from the duplicate threads, in addition to reducing duplication of workload for the forum community \cite{journals/ftir/HoogeveenWBV18}. Each question includes a title, body content and complementary information such as the date of posting and number of votes. It also lists the thread IDs of all duplicates of each thread. The data distribution is highly skewed because only a small fraction (ranging 1.52--9.31\%, depending on the forum) of threads have one or more duplicate.

\subsection{CNET Forum Dataset}

To evaluate the multi-relational classification utility of \DOCDIFFVEC, we use the \cnet forum dataset \cite{W10-2923}, and the dialogue act tagging task (``DA'' hereafter). The dataset is made up of 320 threads comprising 1332 posts from four different subforums of the \cnet\footnote{\url{http://forums.cnet.com/?tag=TOCleftColumn.0}} website. Apart from textual features including post title and body, each post contains structural features such as author name and position of the post in the thread. Each post is manually labelled with one or more parent posts that it relates to, and a unique dialogue act for each link. As the forum is troubleshooting-oriented, the 12 dialogue acts present in the dataset capture the nature of the dialogue interaction, including: \Qq (a newly posed question), \Aa (a solution to a question), and \Qc (correction of an error in a question). In our experiments, we assume knowledge of the parent post(s) of each post, and perform only the DA tagging task. The \cnet dataset has a characteristically skewed class distribution, with the majority DA label (\Aa) accounting for 40.3\% of post pairs in the dataset. 

\subsection{Data Preprocessing}
All textual data in the two datasets is cleaned and tokenized using the script provided with \cqadupstack.\footnote{\url{https://github.com/D1Doris/CQADupStack}} We denote the values (sentence sets) for the two datasets as $S_{dup}$ and $S_{da}$, respectively. Document embedding models are then treated as black box tools that take as input each sentence $s \in \{S_{dup}, S_{da}\} $ and output a vector representation $ \vec{h} \in {\rm I\!R}^{d} $ where $d$ is the dimensionality of the embedding regulated by the sentence encoder.

In order to compare different sentence encoders that generate $\vec{h}$ and result in different \DOCDIFFVEC[s], we use four representative models, two unsupervised and two supervised: word averaging model (unsupervised), \WR model (unsupervised), \infersent (supervised), and \bilstm (supervised). For the unsupervised models, we further enrich the model variety by using different pretrained word embeddings as inputs, including \wordtovec (the 300-dimensional version pre-trained on Google News), \glove and \paragram (the PARAGRAM-SL999 version). In the supervised setting, we use the publicly available Theano implementation\footnote{\url{https://github.com/jwieting/para-nmt-50m}} to train BiLSTM-NMT, and slightly modify the code to convert it into a general-purpose sentence encoder that can vectorize arbitrary text by loading trained models. We preserve all hyperparameters and settings, and use \paragram-SL999 word embeddings\footnote{\url{https://www.cs.cmu.edu/~jwieting/}} to initialize the input sentences according to the original paper \cite{DBLP:conf/acl/GimpelW18}. We train two BiLSTM-NMT models that both output 4096 dimensional document embeddings, with max-pooling and mean pooling, respectively, for comparison. We also keep the native settings for \infersent using the original implementation,\footnote{\url{https://github.com/facebookresearch/InferSent}} where they use \glove word embeddings and a dimensionality of 4096 for output sentence vectors.

For Q-DUP, $r \in R_{cqa} = \{1,0\}$ is a binary variable indicating whether the pair of questions are duplicates or not. Generating all possible pairings $(\vec{h_1}, \vec{h_2}, r)$ for this task leads to an intractable billions of triples, with only a fraction (roughly $1e-6$) being duplicates. For efficiency, we abandon the natural data distribution and choose to keep all duplicates but subsample the non-duplicates to a feasible number, in line with earlier work on the dataset \cite{W16-1609}. Numbers of duplicated pairs range from around 1,000 to 4,000 depending on the subforum. We randomly allocate 90\% of the duplicates to the training set and the other 10\% to the test set for each subforum. We then subsample 5000 times more non-duplicates than duplicates for both training and testing in each subforum.

For DA tagging over \cnet, $r$ belongs to one of the 12 interactive dialogue act tags, e.g. $(\vec{h_1}, \vec{h_2}, \text{\Qq})$. All 1332 labelled training instances are used, and in the instance that a post has multiple parent posts, each is treated as a separate instance. We randomly split the data into 10 folds for cross validation.

We use scikit-learn package in Python to implement the \svm models, with default parameters.

% I illustrate the \DOCDIFFVEC scheme for learning document relations in Figure \ref{fig:docdiffvec}, which turns out to share a similar structure with the Generic NLI training scheme used by Infersent in Figure \ref{fig:joint}.

% \begin{figure}
%     \centering
%     \includegraphics[scale=.25]{docdiffvec}
%     \caption{\DOCDIFFVEC Experiment Scheme}
%     \label{fig:docdiffvec}
% \end{figure}

\begin{table}
\centering
\smaller
\begin{tabular}{lc}
\toprule
  Model & AUC \\
  \midrule
%\multicolumn{14}{|l|}{\emph{Unsupervised Document Embedding Models}} \\ \hline
\Average(\wordtovec) & 0.75 \\
\Average(\glove) & 0.78 \\
\Average(\paragram) & 0.75 \\
\WR(\wordtovec) & 0.74 \\
\WR(\glove) & \textbf{0.79} \\
\WR(\paragram) & 0.75 \\[1.5ex]
%\multicolumn{13}{|l|}{\emph{Supervised Document Embedding Models}} &  \\ \hline

\infersent & \textbf{0.91} \\
\bilstmmax & 0.90 \\
\bilstmavg & 0.85 \\[1.5ex]

%\multicolumn{14}{|l|}{\emph{Previous Works}} \\ \hline
  dbow (\wiki) & \textbf{0.91} \\
  dbow (\AP) & 0.90 \\
  \bottomrule
\end{tabular}%
\caption{AUC Scores for \cqadupstack}
\label{tab:dupstack}
\end{table}

\section{Evaluation and Discussion}

We conduct evaluation from two aspects for the two datasets. In order to evaluate how well \DOCDIFFVEC[s] capture relational differences across the different tasks, we conduct absolute performance comparisons between results produced by \DOCDIFFVEC models and the state-of-the-art models. Our intention here is to determine whether the highly simplistic and general-purpose \DOCDIFFVEC approach is competitive with methods that are customized to the task/dataset. Additionally, we are interested in the cross-comparison  between document embedding models, to determine whether there are substantial empirical differences between them, and the possible causes of any differences.

\subsection{Duplication Detection}
For the Q-DUP task, we train an \svm model for each document embedding method over each of the 12 subforums, and evaluate using the ROC AUC score due to the extremely biased data distribution. The ROC AUC score indicates the probability that the models rank randomly-chosen positive samples before randomly-chosen negative samples. An AUC score of 1.0 indicates that the model is perfect at ranking true duplicates ahead of false duplicates, while 0.5 signifies a completely random ranking (and any value less than that a worse-than-random ranking). As the \svm classifier does not provide an explicit probability to use for ranking, we calculate a similarity score based on:
\begin{equation*}
s_{dup} = \frac{d - d_{\min}}{d_{\max}-d_{\min}}  
\end{equation*}
where $d$ is the distance from the instance to the positive decision boundary of the \svm, and $d_{\min}, d_{\max}$ correspond to the minimum and maximum distances among the test instances. 
We present the AUC results in \tabref{tab:dupstack}.

%This scoring approach has an apparent drawback that for an increasing amount of test data: the range between $d_{min}$ and $d_{max}$ are more prone to be stretched by unexpected outliers in the SVM decision space, causing inaccuracy in the approximated values. Nevertheless, as the overall AUC score is less affected by the values themselves, but relies on the ranking, the impact of this unofficial estimation method is very limited and acceptable. 

For the unsupervised approaches (the top block in the table), the variance between models is slight, but there is a clear pattern that models built on \glove perform slightly better than those built on the other two word embedding models, with the \WR compositional model showing a tiny advantage. While \glove benefits from \WR translation, the other two models do not.

In terms of the more complex supervised models (the middle block), \infersent outperforms \bilstmmax with an very minor advantage and beats the unsupervised models by a large margin. While we only present aggregate numbers in the paper, across all of the individual subforums, max pooling beats mean pooling for the \bilstm model despite all other settings being identical. This could potentially be explained by the phenomenon discussed by \newcite{D17-1070}, that mean pooling does not make sharp enough choices on which part of the sentence is more important. Apart from using the widely adopted BiLSTM architecture for sentence encoding,  the success of the \bilstm model in this task might also benefit from the paraphrastic training objective (optimizing a cosine similarity margin loss). The success of \infersent is not surprising because, in the joint model it computes $(\vec{u}, \vec{v},|\vec{u}-\vec{v}|,\vec{u}\cdot\vec{v})$ as features to predict relations, where \vec{u} and \vec{v} are sentence vectors in a relation pair; that is, it explicitly models $|\vec{u}-\vec{v}|$, which is identical to what we use for \DOCDIFFVEC. Though it is not strictly ``cheating'' as the model is trained on the related but non-identical NLI task, \infersent certainly has the advantage of explicitly capturing \DOCDIFFVEC as a subspace of its larger feature space.  \infersent's small advantage over \bilstm on this task may also be attributable to the different word embeddings it uses (\glove vs.\ \paragram), given that \glove was the pick of the unsupervised methods.
%, or different training corpora and objectives that they have supervision on (parallel text data versus inference data or cosine margin loss versus categorical loss). 

% \begin{figure}
%     \centering
%     \includegraphics[scale=.65]{4}
%     \caption{Impact of Data Distribution}
%     \label{fig:ratio}
% \end{figure}

To calibrate these results against the state of the art for the dataset, we compare ourselves against the best AUC results reported by \newcite{W16-1609}, who fine-tuned \doctovec for Q-DUP. From the different \doctovec models they proposed, we compare ourselves against the best of the \dbow models, which were trained on either English Wikipedia\footnote[1]{Using the dump dated 2015-12-01, cleaned using WikiExtractor: \url{https://github.com/attardi/wikiextractor}} (``\wiki'') or AP-NEWS\footnote[2]{A collection of Associated Press English news articles from 2009 to 2015}  (``\AP'') using pretrained word vectors from \wordtovec. Note that \newcite{W16-1609} simply rank question pairs based on cosine similarity over the \dbow representations. \DOCDIFFVEC[s] obtained from both the \bilstm and \infersent embedding models are highly competitive with the \doctovec approach, with the \infersent model equally \dbow trained on English Wikipedia with an AUC score of 0.91.

%\begin{table*}[]
%\centering
%  \smaller
%\begin{tabular}{c@{\,\,}c@{\,\,}c@{\,}cc@{\,\,}c@{\,\,}ccc@{\,\,}c@{\,\,}cccc}
 % \toprule
 % \multicolumn{7}{c}{Unsupervised Models} && \multicolumn{3}{c}{Supervised Models} && %\multicolumn{2}{c}{Previous Work} \\ 
%  \cmidrule{1-7}
  %\cmidrule{9-11}
  %\cmidrule{13-14}
  %\WR$_{w2v}$ & \WR$_{\glove}$ & \WR$_{\paragram}$ && \Average$_{\wordtovec}$ & \Average$_{\glove}$ & \Average$_{\paragram}$ && \bilstmmax & \bilstmavg & \infersent && \svmhmm & Baseline \\
%  \midrule
 % 0.41 & 0.63 & 0.59 && \textbf{0.65} & 0.63 & 0.60 && 0.57 & \textbf{0.58} & 0.56 && 0.57 & 0.64 \\
 % \bottomrule
%\end{tabular}%
%\caption{Results for \cnet (F-score)}
%\label{rcnet}
%\end{table*}

\begin{table}
\centering
\smaller
\begin{tabular}{llc}
\toprule
  Model & Score \\
  \midrule
%\multicolumn{14}{|l|}{\emph{Unsupervised Document Embedding Models}} \\ \hline
\Average(\wordtovec) & \textbf{0.65} \\
\Average(\glove) & 0.63 \\
\Average(\paragram) & 0.60 \\
\WR(\wordtovec) & 0.41\\
\WR(\glove) & 0.63 \\
\WR(\paragram) & 0.59 \\[1.5ex]
%\multicolumn{13}{|l|}{\emph{Supervised Document Embedding Models}} &  \\ \hline

  \infersent & 0.56 \\
  \bilstmmax & 0.57 \\
  \bilstmavg &\textbf{0.58} \\[1.5ex]

%\multicolumn{14}{|l|}{\emph{Previous Works}} \\ \hline
\svmhmm & 0.57 \\
  baseline & 0.64 \\
  \bottomrule
\end{tabular}%
\caption{Results for \cnet (F1-score)}
\label{rcnet}
\end{table}

\subsection{Dialogue Act Classification}

We test the ability of \DOCDIFFVEC to recognise more complex and diverse relations beyond the binary duplicate detection domain, in the form of the post-to-post dialogue act (DA) tagging task. Here, we evaluate in terms of micro-averaged F1-score. According to \tabref{rcnet}, the unsupervised models (once again, the top block in the table) surprisingly turn the tide to outperform the supervised embedding models, with the simplest averaging model built on \wordtovec attaining an F1-score of 0.65.

In reality, the dialogue act tags depend heavily on structural and contextual features of the post pairs. For example, if we want to detect an answer to a question, the answer is certainly located after where the question is posted, and tends to have a different author to the question requester, as rarely does the requester propose a solution to his/her own question. Similarly, in terms of a \Qc relation, it is likely that the two posts have the same author. Previous approaches using \crf, \svmhmm, \me \cite{W10-2923} and the improved versions using \crfsgd \cite{D11-1002} all make use of such features in addition to words from the post title and body of the post. As \DOCDIFFVEC models do not include those features, we compare them with the \svmhmm models in \citet{W10-2923} that are based solely on lexical features, including lexical unigrams and bigrams, and POS tags. We also compare \DOCDIFFVEC models with the heuristic baseline of \citet{W10-2923}, where the first post is always classified as a \Qq, and all subsequent posts are classified as an \Aa. This baseline achieved a reasonably high F1-score of 0.64, due to the high proportion of \Aa and \Qq tags, and the utility of positional information. Our best \DOCDIFFVEC model passes the baseline by a mere 0.01, but comfortably beats the \svmhmm model. It was unexpected that all supervised models would perform poorly, below the baseline by quite a margin, at a similar level to the \svmhmm model.

Note that the state of the art result for this dataset is that of \citet{I17-1056} based on a memory-augmented CRF, with structural and post author features as side information. They achieve an F1-score of 0.78 with a much more complex supervised model, clearly above our best result, but given the simplicity and flexibility of our approach, an F-score of 0.65 is plausibly competitive.

By analyzing the confusion matrix for the \Average(\wordtovec) model, we found that only the \Aa, \Qq and \Rr relations are correctly recognized at an acceptable level, which are the three most common tags in the data. For rarer tags, the F1-scores approach 0, indicating that the model has limited ability to further distinguish \DOCDIFFVEC[s] into more specific subclasses. Also, the three well-classified tags have recall greater than precision, which suggests a tendency for the model to classify unknown or uncertain relations into dominant classes. As most \Qq relations are associated with reentrant links (the link from the parent node in the thread is to itself), the \DOCDIFFVEC will always be \vec{0}, which is easy for the classifiers to detect, leading to a particularly high F1-score for the class of 0.9. 

It is interesting that for the \bilstm models, precision is quite a bit lower than recall for the \Qq tag, which seems to be the main reason for their poor performance. Tying the result back to Q-DUP, supervised embedding models suffer from their unnecessarily strong ability to capture semantic similarity in their \DOCDIFFVEC[s] for the DA task. Most linked posts in this dataset are very similar in terms of topic and content, as they belong to the same question thread and discuss the same specific issue. This causes posts to have high semantic similarity within a thread, and the \DOCDIFFVEC[s] to be close to 0 in magnitude in vector space. This causes the prevalence of \Qq misclassifications.

More generally, the reason why \DOCDIFFVEC models do not perform better can be ascribed to: (1) the \svm overfitting to the majority tags, due to data sparsity; (2) subtle distinctions between less common dialogue acts being difficult to make without structural features or post metadata (e.g.\ author, position of post), regardless of the document embedding model used; and (3) \DOCDIFFVEC being incapable of differentiating multiple dialogue acts in a single linear vector space. 

Overall, we cautiously conclude that \DOCDIFFVEC[s] have quantifiable but ultimately limited utility for multi-relational classification, especially in contexts where extra-linguistic factors have high import.

\subsection{Discussion and Future Work}

Drawing the two tasks together, we can observe that \DOCDIFFVEC[s] have remarkable utility in document similarity modelling, but are weaker at multi-relational classification tasks. Ultimately, however, further experimentation over other tasks is required to determine how well \DOCDIFFVEC performs over tasks beyond document similarity, such as entailment, summarization (e.g.\ body of an article versus its title), and question-answering.

The conclusion that \DOCDIFFVEC does not model multi-relational classification tasks well ties in with recent work on knowledge graph (``KG'') embedding, such as the \transr model \cite{conf/aaai/LinLSLZ15}. Traditional KG embedding models represent all relations and entities in a single semantic space, regarding a relation as a translation from a head entity to a tail entity (similar to \DIFFVEC). Nevertheless, one semantic space is considered insufficient because each pair of entities is likely to be associated across a number of relations. To overcome the multi-relational weakness of KG embedding \DIFFVEC[s], \transr learns an exclusive vector space for each distinct relation, and shows this to result in significant improvement for KG embedding. The relations in \transr are still represented in \DIFFVEC form but are calculated after transforming entity embeddings into a vector space customized to a given relation. Familiarly enough, when training universal document embedding models, sentences are trained to be ``entities'' in a semantic space. In future work, we propose to assess whether \DOCDIFFVEC[s] can automatically capture relations between sentences during training, despite not being explicitly trained to learn relations, as in KG models. That is, the approach of \transr is potentially also a good fix for multi-relational learning at the document level.

Finally, instead of using a simple linear kernel \svm to classify \DOCDIFFVEC[s], it would of course be possible to use more sophisticated classifiers, such as deep neural networks, to explore more complex, non-linear composition of the directions and magnitudes encoded in \DOCDIFFVEC[s].

\section{Conclusions}

Taking inspiration from the work of word-level embedding vector offsets for lexical relation learning, this paper is the first to evaluate document embedding vector difference vectors to model document-to-document relations. By using a simple \svm model to classify \DOCDIFFVEC[s], we found that BiLSTM-based document embedding models generate \DOCDIFFVEC[s] that are highly useful for textual similarity, through experiments on a document duplication detection task, competitive with the state of the art. At the same time, we found that \DOCDIFFVEC[s] obtained from simple averaging of word embeddings outperform an informed baseline and complex neural sentence encoders for multi-relational classification, in the context of dialogue act tagging. Overall, we conclude that \DOCDIFFVEC has reasonable utility for document relation learning.

%%%%%%%%%%%%%%%%%%%%%%%%%%%%%%%%%%%%%%%%%%%%%%%%%%%%%%%%%%%%%%%%%%%%%%%%%%%%%%%%

%%%%%%%%%%%%%%%%%%%%%%%%%%%%%%%%%%%%%%%%%%%%%%%%%%%%%%%%%%%%%%%%%%%%%%%%%%%%%%%%

%%%%%%%%%%%%%%%%%%%%%%%%%%%%%%%%%%%%%%%%%%%%%%%%%%%%%%%%%%%%%%%%%%%%%%%%%%%%%%%%

% \section{ACKNOWLEDGMENT}
% I appreciate authors of Scikit-Learn, nltk, python, Theano and PyTorch for great assistance and convenience brought to the implementation of experiments.

\bibliographystyle{acl_natbib}
\bibliography{main}

\end{document}